\documentclass[english,12pt]{article}
\usepackage[T1]{fontenc}
\usepackage[latin9]{inputenc}
\usepackage{bm}
\usepackage{amssymb}
\usepackage{amsmath}
\usepackage{xcolor}
\usepackage[margin=1.0in,letterpaper]{geometry}
\usepackage{algorithmic}
\usepackage{tikz}
\usepackage{color}
\usepackage{multirow}
\usetikzlibrary{bayesnet}
\usetikzlibrary{arrows}
\usetikzlibrary{positioning}
\usetikzlibrary{backgrounds}
\usepackage[unicode=true,pdfusetitle,
bookmarks=true,bookmarksnumbered=false,bookmarksopen=false,
breaklinks=false,pdfborder={0 0 1},backref=false,colorlinks=false]{hyperref}
\makeatletter
\usepackage{wasysym}
\usepackage{algorithm}
\usepackage{algorithmic}

\usepackage{bbm}
\usepackage{babel}
\date{}

\makeatother

\begin{document}
\title{Contrastive Enhancement Using Latent Prototype for Few-Shot Segmentation}
\author{
	Xiaoyu Zhao, Xiaoqian Chen, Zhiqiang Gong, \\ 
	Wen Yao, Yunyang Zhang, Xiaohu Zheng, \\
	zhaoxiaoyu13@nudt.edu.cn
}
\maketitle

\begin{abstract}
	Few-shot segmentation enables the model to recognize unseen classes with few annotated examples. Most existing methods adopt prototype learning architecture, where support prototype vectors are expanded and concatenated with query features to perform conditional segmentation. However, such framework potentially focuses more on query features while may neglect the similarity between support and query features. This paper proposes a contrastive enhancement approach using latent prototypes to leverage latent classes and raise the utilization of similarity information between prototype and query features. Specifically, a latent prototype sampling module is proposed to generate pseudo-mask and novel prototypes based on features similarity. The module conveniently conducts end-to-end learning and has no strong dependence on clustering numbers like cluster-based method. Besides, a contrastive enhancement module is developed to drive models to provide different predictions with the same query features. Our method can be used as an auxiliary module to flexibly integrate into other baselines for a better segmentation performance. Extensive experiments show our approach remarkably improves the performance of state-of-the-art methods for 1-shot and 5-shot segmentation, especially outperforming baseline by 5.9\% and 7.3\% for 5-shot task on Pascal-$5^i$ and COCO-$20^i$. Source code is available at \url{https://github.com/zhaoxiaoyu1995/CELP-Pytorch} \\ \\
	\noindent \textbf{Keywords:} Few-shot Segmentation, Latent Prototype, Contrastive Enhancement
\end{abstract}

\section{Introduction}

Deep learning-based segmentation methods have achieved state-of-the-art in various image segmentation tasks, benefiting from large pixel-level annotated datasets and advances in deep neural networks \cite{45-gong2020statistical,44-gong2019cnn}. 
However, available labeled samples for the segmentation task are usually limited since pixel-level annotation is expensive and time-consuming.
Specifically, the performance of fully-supervised approaches drops dramatically for tasks without sufficient annotated data.
Though semi-supervised methods succeed in leveraging unlabeled samples as the complement of annotated data, segmentation assisting with a few or even one annotated sample is extremely difficult.
In particular, these methods usually cannot be generalized to unseen classes.

Few-shot segmentation \cite{32-yang2021mining,33-zhang2020sg,34-zhang2019canet,12-tian2020prior} aims to segment novel objects in a query image using few annotated support images.
Existing methods mostly leverage representations from annotated support images and the similarity between the representations and support features for dense predictions. 
Additionally, prototype learning architecture is a milestone, which extracts single or multi-prototype vectors from support images to represent object information. For example, masked Global Average Pooling (GAP) \cite{42-zhou2016learning} is mostly performed to extract prototypes from support features. 
Some methods adopt EM algorithm and cluster to generate multi-prototype to represent more support information. 
Then the similarity between the prototype vectors and query features is computed using cosine distance \cite{33-zhang2020sg} or dense comparison \cite{34-zhang2019canet} for segmentation.

\begin{figure}
	\centering
	\includegraphics[width=1.0\linewidth]{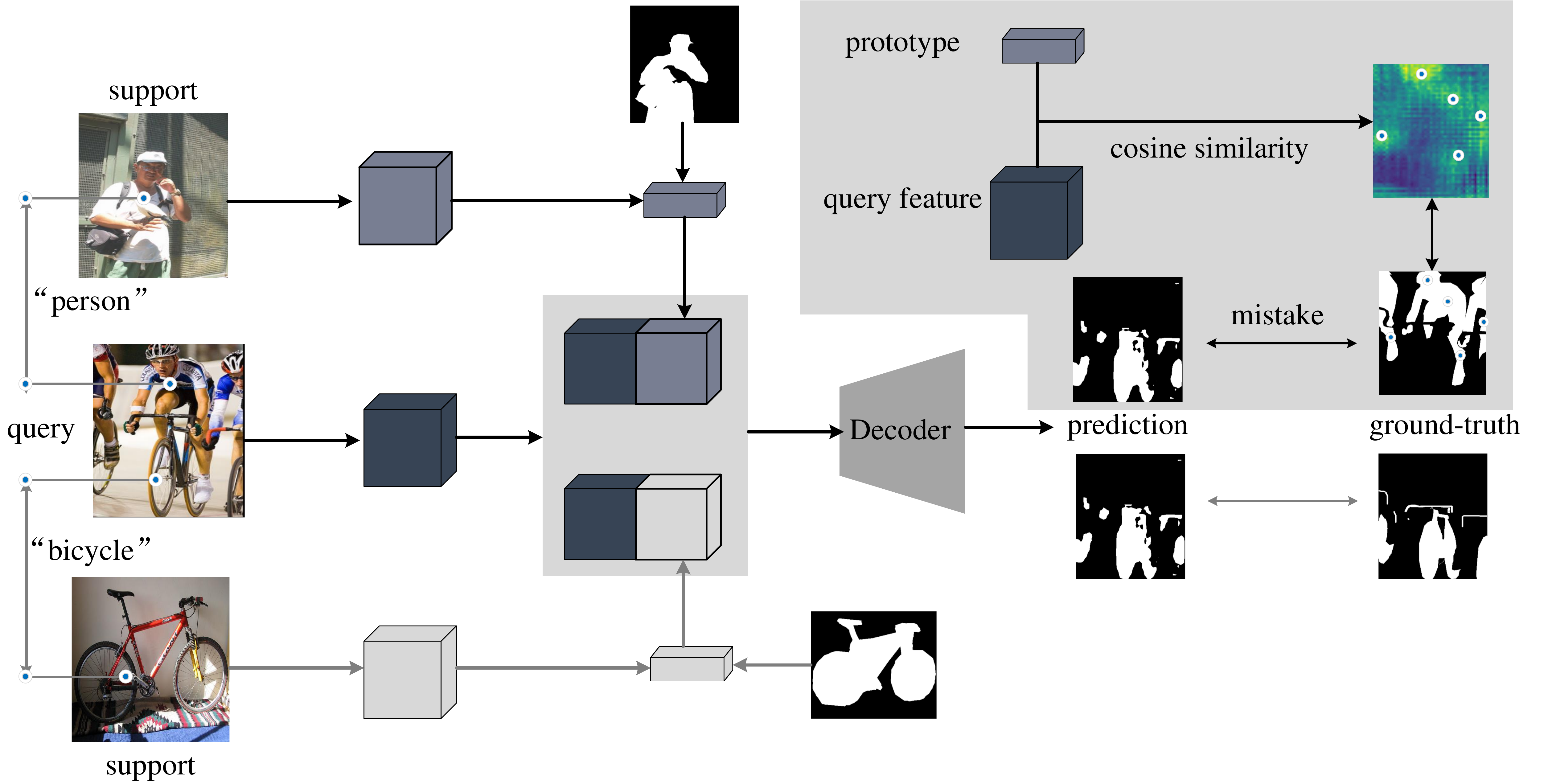}
	\caption{The motivation of our method. 
		In the prototype learning architecture with conditional segmentation, decoder may neglect the similarity between prototype and query features, and overly concentrate on query features for highly accurate prediction in known categories. 
		It may cause segmentation on incorrect objects. 
		Our approach aims to mine latent prototypes and enhance the model to make different predictions with various prototypes.}
	\label{fig:motivation}
\end{figure}

As a learnable pattern, conditional segmentation by dense comparison is generally superior to cosine distance in prediction accuracy.
However, we discover that conditional segmentation sometimes neglects the similarity between prototype and query features, causing incorrect objects segmentation. 
As shown in Fig.~\ref{fig:motivation}, the support and query images contain the objects of ``Person'' and ``Bicycle'', and ``Person'' is unseen and novel to the model. 
Though the prototype represented ``Person'' is extracted by GAP for query, the decoder mistakenly segments the object of ``Bicycle''. However, there exists a high similarity between prototype vector and query features in the ``Person'' object regions.
We suppose that decoder overly concentrates on query features for highly accurate prediction in known categories, neglecting the similarity between prototype and query features, especially when the unseen category has a larger difference with existing categories. 
Some prior works have tried to address such a problem.
PFENet \cite{12-tian2020prior} proposed a training-free prior mask generation method to introduce high-level semantic information. 
CyCTR \cite{23-zhang2021few} designed a Transformer architecture with a cycle-consistent mechanism to aggregate pixel-wise support features.
HSNet \cite{20-min2021hypercorrelation} leverages multi-level feature correlation between support and query feature instead of the prototype learning architecture.
However, these methods do not explicitly utilize background features and constraint conditional segmentation to avoid inconsistent object segmentation.

In this paper, we propose a contrastive approach with latent prototype to enhance decoder in conditional segmentation. 
Especially, we design a Latent Prototype Sampling (LPS) module to generate prototypes from background in query image. 
Our method is based on the hypothesis that high similarity regions in high-level features belong to the same category. 
Though the regions may not represent the whole object, decoder should segment part of objects employing the generated prototype from the sampling regions and suppress the known categories (e.g. the ``Bicycle'' in Fig. \ref{fig:motivation}). 
Moreover, we design a Contrastive Enhancement (CE) module, which can be used as an auxiliary module to leverage latent prototypes to segment corresponding regions.
Our module does not need extra parameters and is transferable since it can share parameters with the decoder module and can be easily applied in other methods to improve segmentation performance further.
Extensive experiments show that our method significantly improve the performance of existing baseline methods. 

In summary, the contributions of our work are as follows:
\begin{itemize}
	\item[$\bullet$] We propose a latent prototype sampling module to generate prototypes from background of query images.
	The module can mine the latent objects in the training images and produce novel prototypes from query features, which is directly utilized in contrastive enhancement of conditional segmentation.
	
	\item[$\bullet$]  We propose a contrastive enhancement module to reduce the over-concentration on query features and the neglect of feature similarity in dense comparison. The module is an auxiliary path of the existing decoder without additional parameters.
	
	\item[$\bullet$] Our method is transferable and can be conveniently applied to existing methods. Experiments show that our approach achieves state-of-the-art performance on few-shot segmentation datasets. The proposed method respectively obtains mIoUs of 65.8\%/40.2\% for 1-shot, and mIoUs of 67.3\%/45.3\% for 5-shot on Pascal-$5^i$ and COCO-$20^i$.
\end{itemize}

\section{Related Works}

\subsection{Fully-Supervised Semantic Segmentation}

Semantic segmentation is a challenging computer vision task to provide pixel-level predictions. 
Fully Convolutional Networks (FCNs) \cite{1-long2015fully} replaced fully connected layers with convolution layers, which is a milestone to promote end-to-end semantic segmentation with deep learning. Based on FCNs architecture, some semantic segmentation methods focus on multi-scale feature \cite{3-chen2017deeplab,6-ronneberger2015u,7-he2019adaptive,8-zhao2017pyramid}, large receptive field \cite{2-chen2014semantic,4-chen2017rethinking} and high-resolution predictions \cite{9-badrinarayanan2017segnet,10-fourure2017residual}. 
Typically, DeepLab methods \cite{2-chen2014semantic,3-chen2017deeplab,4-chen2017rethinking,5-chen2018encoder} proposed dilated convolutions to obtain large receptive field and boost feature resolution in downsampling. In addition, Atrous Spatial Pyramid Pooling (ASPP) module is designed to aggregate multi-scale features. 
Different from FCNs, recent research shows transformer architectures \cite{11-liu2021swin} can also achieve state-of-the-art performance in semantic segmentation. 
Although these methods significantly improve the prediction accuracy of semantic segmentation, they need expensive and time-consuming pixel-level annotations for amounts of images. It is challenging for these methods when few labeled images can be obtained.

\subsection{Few-Shot Learning}

Few-shot learning aims to improve generalization performance of model with few labeled samples. The main research focuses on the data, model, and algorithm perspectives \cite{43-wang2020generalizing}. 
From perspective of data, data augmentation \cite{24-zhang2017mixup,25-yun2019cutmix} generates diverse training samples to promote training, which is also a common technology in fully-supervised learning.
Meta-learning \cite{27-finn2017model,26-li2017meta} is the representative algorithm-based method, targeting to learn from various tasks for better generalization in the new task.
Metric learning methods are important research lines from the model perspective. For the few-shot classification task, Prototypical Network \cite{28-snell2017prototypical} calculated similarity with prototype representation of each class, and highest similarity represents the same pairs. This kind of method focuses on better prototype generation and similarity measurement. 
Matching Network \cite{30-vinyals2016matching} utilized external memory to store knowledge related to tasks and adopts an attention mechanism to read and update memory modules. Using cosine similarity as distance criterion, BD-CSPN \cite{29-liu2020prototype} proposed label propagation and feature shifting to diminish the intra-class and inter-class bias.
Our work follows the research of metric learning, specifically the prototype learning framework in few-shot segmentation task. Few-shot segmentation is a pixel-level classification task with very few examples, which have gained extensive attention recently. 

\subsection{Few-Shot Segmentation}

Few-Shot Segmentation methods are mainly based on metric learning and extend from few-shot image classification methods. 
\cite{31-dong2018few} early formulated n-way k-shot few-shot semantic segmentation and introduced the prototype learning framework. Most methods follow the idea that segmentation is provided according to the similarity between prototypes from support images and pixels in query images. 
SG-One \cite{33-zhang2020sg} proposed masked GAP to generate prototypes from support images and adopted cosine similarity to perform prediction. 
CANet \cite{34-zhang2019canet} replaced cosine similarity with dense comparison, which upsamples the prototype vectors and concatenates them with query features to make conditional segmentation.
Some works pointed out that multiple prototypes can represent more information than single prototype and proposed multiple prototype extraction methods, including expectation-maximization algorithm \cite{35-yang2020prototype}, superpixel-guided clustering \cite{7-he2019adaptive}, and self-guided learning \cite{13-zhang2021self}.
Leveraging information of high-level semantic and background regions is an important research trend.
PFENet \cite{12-tian2020prior} generated prior masks from high-level features to guide predictions. To utilize backround region of training images, \cite{32-yang2021mining} proposed latent class mining strategy and rectification method for support prototypes. CyCTR \cite{23-zhang2021few} proposed cycle-consistent mechanism and integrated it into Transformer architecture, aiming to use the information of whole support features.    
Some works \cite{36-wang2020few,20-min2021hypercorrelation} directly adopted pixel-to-pixel feature correlation between support and query images to make predictions.
This paper proposes a contrastive enhancement method to leverage the high-level feature correlation and background query pixel features. The method enables to discover latent prototypes and regularize conditional segmentation. Besides, it is a flexible module to integrate into existing prototype-based methods, such as PFENet and CyCTR. 

\section{Problem Setting}
Few-shot segmentation aims to train a model to segment unseen objects with few labeled images of the target category. Annotated images for these categories do not appear in the training dataset. 
Given the training dataset ${D_{{\rm{train}}}}$ with classes set ${C_{{\rm{train}}}}$, and test dataset ${D_{{\rm{test}}}}$ with classes set  ${C_{{\rm{test}}}}$, test set is not overlap with train set, i.e. ${C_{{\rm{train}}}} \cap {C_{{\rm{test}}}} = \emptyset $. Consistent with previous works \cite{12-tian2020prior,34-zhang2019canet,35-yang2020prototype}, we preform episode training for few-shot segmentation. 
For K-shot segmentation, each episode is composed with a support set ${S_{{\rm{train}}}} = \{ ({I_S},{M_S})\}$ and a query set ${Q_{{\rm{train}}}} = \{ ({I_Q},{M_Q})\} $, where $I \in {\mathbb{R}^{H \times W \times 3}}$ denotes the RGB image and $M \in {\mathbb{R}^{H \times W}}$ represents mask. 
Images in the support set and query set contain the same category objects $c^*$ annotated in the segmentation mask. In each episode, the model learns to segment class $c^*$ in query images using the support images and support masks. Support-query paradigm is also conducted during testing, but only support masks are available for model input.

\section{Methodology}

\subsection{Overview}

\begin{figure}
	\centering
	\includegraphics[width=1.0\linewidth]{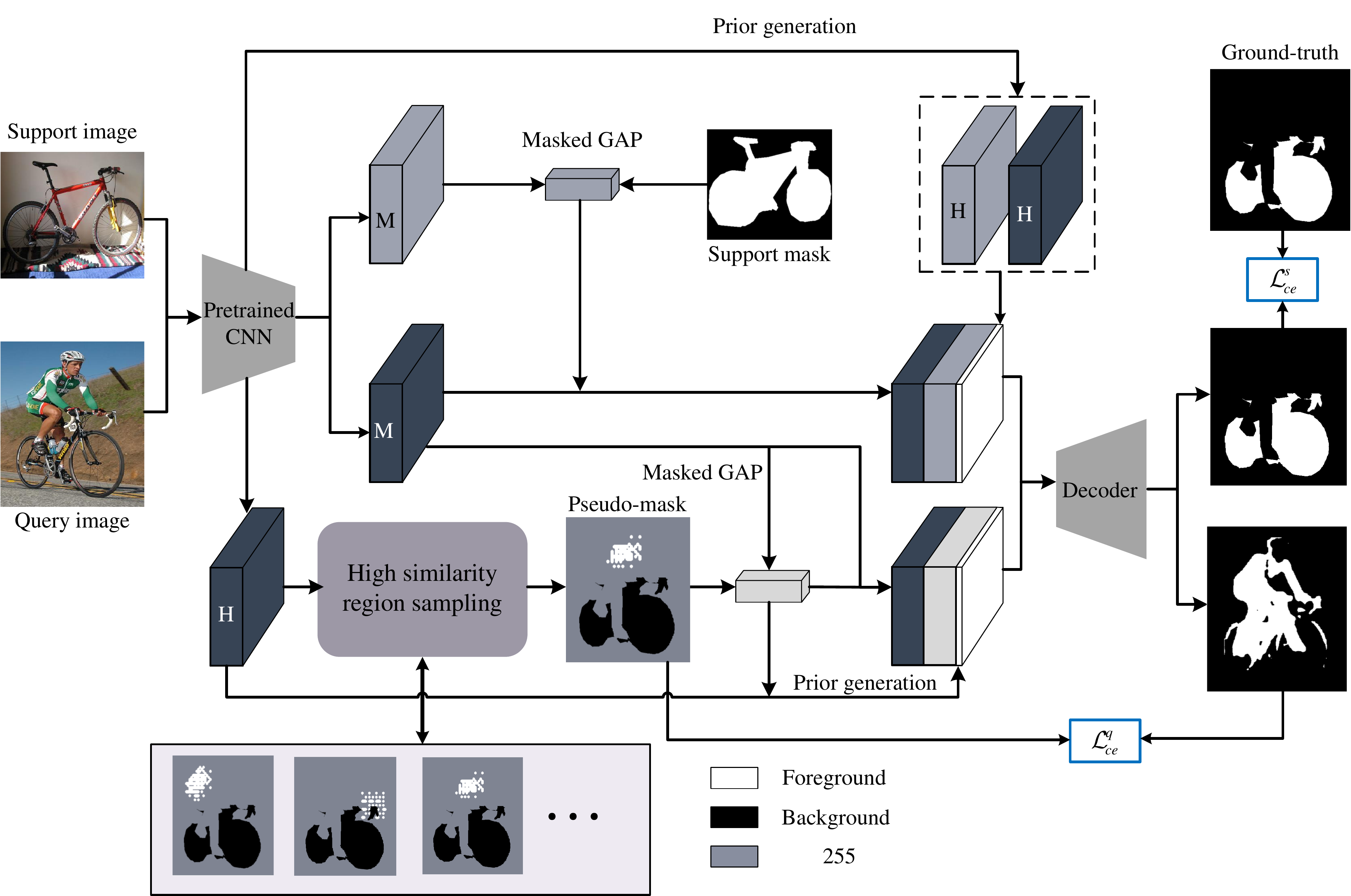}
	\caption{The pipeline of our method  for 1-shot segmentation. M and H features represent the middle and last block outputs of backbone. The top part is the main path, which can be the prototype-based baseline methods. In the auxiliary path, query image regions with high similarity in high-level features are sampled to generate latent prototypes. The latent and support prototypes are respectively concatenated with the same query feature to make different predictions. The additional path enables to mine novel classes and constraint the model training.}
	\label{fig:pipline}
\end{figure}

Fig.~\ref{fig:pipline} shows the overview of our proposed contrastive enhancement method using latent prototype. We adopt the episodic training framework on support-query pairs. 

Initially, the framework employs a pre-trained backbone to extract the middle and high-level features of support and query images, which are generally the outputs of the middle and last blocks in VGG and ResNet. Following \cite{12-tian2020prior}, the similarity between high-level support and query feature map is computed as prior masks to assist predictions. The middle-level support feature maps generate prototype vectors, which are expanded and concatenated with query feature maps and prior masks. The decoder, for instance FEM module \cite{12-tian2020prior} and cycle-consistent transformer \cite{23-zhang2021few}, is to make dense predictions.

Furthermore, parallel to the above process, our framework consists of two modules: Latent Prototype Sampling (LPS) and Contrastive Enhancement (CE) modules. LPS module is designed to obtain the regions that belong to the same category in query images. The sampling strategy 
is established that regions of the same category have high similarity in high-level feature maps. The masked GAP and the generated pseudo-mask are employed to produce latent prototype. 

Finally, CE module computes the similarity between latent prototype and middle-level query feature as prior mask in auxiliary path. The prior mask, latent prototype, and query feature are given to the decoder to predict the sampling regions, where the known objects as background and sampling regions as foreground. This module drives the same query feature combined with different prototype vectors segmenting different objects.

\subsection{Latent Prototype Sampling}

Latent Prototype Sampling (LPS) module is proposed to mine latent stuff, some of which are the unseen foreground objects in labeled data. Supposing the query image $I_q$ and query mask $M_q$, the middle-level and high-level features extracted from the backbone are denoted $F_q^m$ and $F_q^h$. The cosine similarity between feature vectors at position $i$ and $j$ in high-level feature map $F_q^h$ is calculated by
\begin{equation}
	{D_q}(i,j) = \cos (F_q^h(i),F_q^h(j)) = \frac{{F_q^h{{(i)}^{\rm{T}}}F_q^h(j)}}{{\left\| {F_q^h(i)} \right\|\left\| {F_q^h(j)} \right\|}},i,j \in \{ 1,2, \ldots ,hw\},
\end{equation}
where $h$ and $w$ are the height and width of the feature map $F_q^h$. For each position $i$ of feature $F_q^h(i)$, the total number of feature vectors that has higher similarity with $F_q^h(i)$ is recorded. Significantly, the feature that belongs to the foreground annotated in query mask $M_q$ is ignored because the module focus on mining unknown objects. The criterion of higher similarity is controlled through hyper-parameter $\delta$ by
\begin{equation}
	N(i) = \sum\nolimits_j^{hw} {\left[ {{D_q}(i,j) >  = \delta } \right] \cdot } \left[ {{M_q}(j) = 0} \right]~~~i \in \{ 1,2, \ldots ,hw\}.
\end{equation}

We expect to sample regions based on the central point, which is distinctly similar to surrounding points. Then, the index set of the central feature vector can be constructed as follow:
\begin{equation}
	P = \left\{ {i\left| {N(i) \ge \sigma ,{M_q}(i) = 0,i \in \{ 1,2,...,hw\} } \right.} \right\},
\end{equation}
where $\sigma$ is a threshold to filter out isolated points with very few similar features. The chosen index $i^*$ is randomly selected from set $P$. The regions containing $i^*$ and $i^*$'s similar features are supposed to be the same class. The known foreground in annotated query mask belongs to different categories with the sampling regions and is treated as background. Other regions that are not identified as the background and foreground are ignored in loss computation to avoid introducing mistakes. The mask generation is described as follow:
\begin{equation}
	{M_q}^1 = \left[ {{D_q(i^*,\cdot)} >  = \delta } \right]{M_q}^{\rm{T}},
\end{equation}
\begin{equation}
	{M_q}^{255} = 255 \cdot [{M_q}^1 =  = 0] \cdot [{M_q} == 0],
\end{equation}
\begin{equation}
	{M_q}^\prime  = {M_q}^1 + {M_q}^{255},
\end{equation}
where ${M_q}^1$, ${M_q}^{255}$, and ${M_q}^\prime$ are respectively the foreground, ignore regions and generated mask. The sampling procedure can be conveniently conducted by matrix operations, hence the increase in computing time is negligible. We design the clear sampling methods rather than the existing clustering method because the numbers of clustering centers are hard to determine, especially for each query image. Unreliable clustering will bring label noise and damage the original training branch. Our approach is simple for implementation and efficient to conduct end-to-end training.
The masked GAP takes the middle-level features $F_q^m$ and the generated mask ${M_q}^\prime$ to produce the latent prototype $v_l$, i.e.
\begin{equation}
	{v_l} = \frac{{\sum\nolimits_{i = 1}^{hw} {F_q^m \cdot \left[ {{M_q}^\prime (i) = 1} \right]} }}{{\sum\nolimits_{i = 1}^{hw} {\left[ {{M_q}^\prime (i) = 1} \right]} }}.
\end{equation}  

The latent prototype and generated pseudo-mask are prepared for the contrastive enhancement module. 

\subsection{Contrastive Enhancement}

Contrastive Enhancement (CE) is a parallel path to enhance activation in unknown classes and increase attention in high similarity parts. The similarities between prototype vectors and high-level features are firstly computed to generate a prior mask ${H_q}^\prime$, which tells the probability of being the same categories:
\begin{equation}
	{H_q}^\prime  = \cos ({v_l},F_q^m \odot {M_q}^\prime ),
\end{equation}
where $\odot$ is the Hadamard product, which sets the background in feature map to zeros. Besides, the values in ${H_q}^\prime$ are normalized to 0 and 1 as
\begin{equation}
	{H_q}^\prime  = \frac{{{H_q}^\prime  - \min ({H_q}^\prime )}}{{\max ({H_q}^\prime ) - \min ({H_q}^\prime ) + \varepsilon }},
\end{equation} 
where $\varepsilon$ is a small value for numerical stability. Using dense comparison, the latent prototype $v_l$ is expanded to $V_l$, which is concatenated with query feature $F_q^m$ and prior mask ${H_q}^\prime$ as the input feature maps ${F_q}^\prime$ of decoder:
\begin{equation}
	{F_q}^\prime  = Concat\left( {\left[ {F_q^m,{V_l},{H_q}^\prime } \right]} \right).
\end{equation}

In the main path, the support prototype ${v_s}$ is generated using support mask $M_s$ and support feature $F_s^m$:
\begin{equation}
	{v_s} = \frac{{\sum\nolimits_{i = 1}^{hw} {F_s^m \cdot \left[ {M_s (i) = 1} \right]} }}{{\sum\nolimits_{i = 1}^{hw} {\left[ {M_s (i) = 1} \right]} }}.
\end{equation}

We generate the prior mask between support features and query features following \cite{12-tian2020prior}. For each position $i$, ${Y_q}(i)$ represents the maximum similarity among each query feature $F_q^h(i)$ and all support features $F_s^h$. $H_q$ is the min-max normalization of $Y_q$. Denoting $V_s$ as the expanded support prototype, the input $F_q$ of encoder in the main path is described as follows: 
\begin{equation}
	F_q  = Concat\left( {\left[ {F_q^m,{V_s},H_q } \right]} \right).
\end{equation}
\begin{figure}
	\centering
	\includegraphics[width=1.0\linewidth]{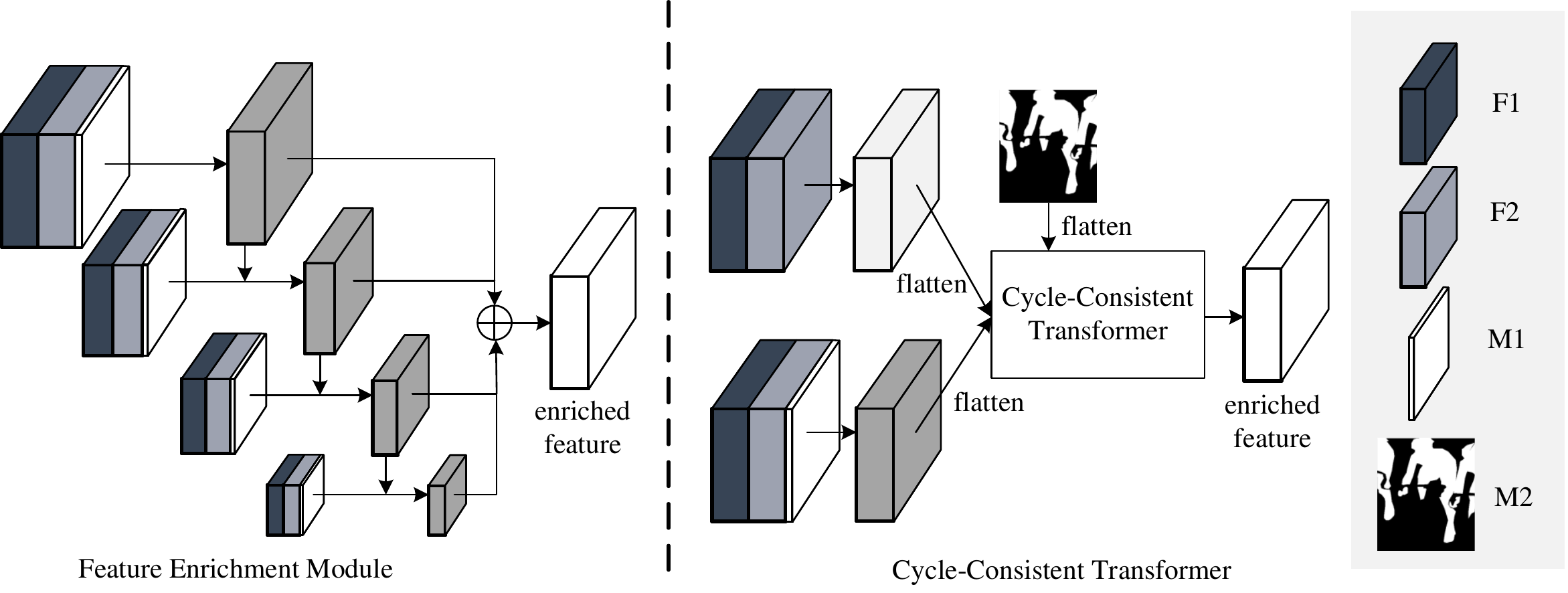}
	\caption{The architecture of FEM \cite{12-tian2020prior} and CCT \cite{23-zhang2021few}. In our CE module, F1 is the query features. In the main path, M2, F2 are respectively query mask and expanded support prototype. M1 is the prior mask between support features and query features. In the auxiliary path, M2 and F2 are pseudo-mask and expanded latent prototype. M1 is the prior mask between latent prototype and query features.}
	\label{fig:decoder}
\end{figure}

For the predicted probability maps in two paths, the segmentation of query images is supervised by the cross-entropy loss. The overall training loss consists of three parts as follows:
\begin{equation}
	{\cal L} = {{\rm{{\cal L}}}_{ce}^{1}}\left( {{P_q},{M_q}} \right) + \alpha {{\rm{{\cal L}}}_{ce}^{2}}\left( {{P_q}^\prime ,{M_q}^\prime} \right) + \beta{\rm{{\cal L}}}_{ce}^{3},
\end{equation}
where ${{\rm{{\cal L}}}}\left(  \cdot  \right)$ is the cross-entropy loss function. Additionally, ${\rm{{\cal L}}}_{ce}^{2}$ is the proposed contrastive enhancement loss, while ${\rm{{\cal L}}}_{ce}^{3}$ is the specific loss in various decoders. $P_q$ and ${{P_q}^\prime}$ are respectively the predicted probability maps of query mask $M_q$ and pseudo-mask ${M_q}^\prime$. $\alpha$ and $\beta$ are the weight to balance the effect of multi-losses.

CE module can be adapted to various architecture of decoder. In our work, the Feature Enrichment Module (FEM) \cite{12-tian2020prior} and Cycle-Consistent Transformer (CCT) \cite{23-zhang2021few} are respectively used to obtain enriched query feature. As shown in Fig.~\ref{fig:decoder}, FEM is a multi-scale structure to leverage multi-level spatial information. For the multi-scale prediction $P_q^i\left\{ {i = 1,2, \ldots ,n} \right\}$. The auxiliary loss ${\rm{{\cal L}}}_{3}$ is defined as
\begin{equation}
	{\rm{{\cal L}}}_{ce}^{3} = \sum\nolimits_{i = 1}^n {{{\rm{{\cal L}}}_{ce}}\left( {P_q^i,M_q^i} \right)}.
\end{equation}

CCT consists of two transformer blocks, where the self-alignment block aggregates global context within query features $F_q$, and the cross-alignment block leverages the information in the support features $F_s$. In this architecture, the predicted query mask is used to segmentation the support images to constraint the model training. Denoting $P_s$ as the predicted support mask ${M_s}$, the loss ${\rm{{\cal L}}}_{3}$ can be described as
\begin{equation}
	{\rm{{\cal L}}}_{ce}^{3} = {{\rm{{\cal L}}}_{ce}}\left( {{P_s},{M_s}} \right).
\end{equation}

\subsection{K-Shot Segmentation}
\label{sec:k-shot}
For K-shot setting, the model is trained on 1-shot task and directly used for 5-shot segmentation evaluation. The support prototype vector $v_s$ and prior mask $H_q$ are respectively the average of $\left\{ {v_s^i} \right\}_{i = 1}^k$ and $\left\{ {H_q^k} \right\}_{i = 1}^k$ from K support images.

\section{Experiments}

\subsection{Datasets and Evaluation Metric}
Following previous studies \cite{12-tian2020prior,13-zhang2021self,14-li2021adaptive}, we evaluate our methods on two widely used datasets, including Pascal-$5^i$ and COCO-$20^i$. 
Pascal-$5^i$ is combined with PASCAL VOC 2012 \cite{37-everingham2010pascal} and SBD \cite{38-hariharan2014simultaneous} dataset, in which 20 classes are divided into 4 splits. In each experiment, 3 splits are chosen for training and the rest for testing. During evaluation, 1000 support-query pairs from each class are sampled. 
COCO-$20^i$ is larger and more complex than Pascal-$5^i$. It is built based on MS-COCO \cite{39-lin2014microsoft} dataset, which contains 80 classes and 82081 images in training set. Similar to Pascal-$5^i$, 80 classes are split into 4 folds for cross-validation, but total of 20000 support-query pairs are randomly sampled for reliable evaluation \cite{12-tian2020prior}. 
Besides, the mean intersection over union (mIoU) and foreground-background intersection-over-union (FB-IoU), the much common evaluation metric in segmentation task, are adopted in contrast experiment and ablation studies.

\subsection{Implementation Details}
Since our approach is flexible to integrated into other few-shot segmentation approachs, we implement different versions based on two baselines: PFENet \cite{12-tian2020prior} and CyCTR \cite{23-zhang2021few}.  
VGG16, ResNet50, and ResNet101 pre-trained on ImageNet are the backbone to extract features. The feature map in block4 is used to generate pseudo-mask for contrastive enhancement. Regions with high similarity in query features are randomly selected as foreground masks, and masked GAP is adopted to generate latent prototype vectors. 
The similarity threshold is set to 0.65. All backbone parameters are frozen, including batchnorm. 

For fair comparisons, Images are resized and cropped to 473x473 (Pascal) or 641x641 (COCO) for PFENet implementation, while the sizes are both 473x473 for CyCTR implementation. 
The data augmentation is the same as previous methods, including random rotation, gaussian blur, horizontal flip, and crop. Dice loss \cite{41-milletari2016v} is used as loss function in CyCTR and cross-entropy in PFENet. 
The total epochs are set to 50 for COCO-$20^i$ and 200 for Pascal-$5^i$. 
For transformer blocks in CyCTR, AdamW \cite{40-loshchilov2017decoupled} optimizer is used with initial learning rate $1 \times 10^{-4}$ and weight decay $1 \times 10^{-2}$. Other blocks are optimized by SGD with initial learning rate $2.5 \times 10^{-3}$ for Pascal-$5^i$ and $5 \times 10^{-3}$ for COCO-$20^i$. 
The hyperparameter of weight $\alpha$ and $\beta$ are set to 1.0 and 0.1.
The poly learning rate decay with 0.9 power is used. The mini-batch size is set to 4 for Pascal-$5^i$ and 8 for COCO-$20^i$. We implement our methods using Pytorch, and all experiments are conducted on single NVIDIA GeForce RTX 3090 GPU.

\subsection{Comparisons with State-of-the-Art}

\begin{table}[htb]
	\centering
	\caption{Comparison with other state-of-the-arts using mIoU(\%) for 1-shot and 5-shot settings on Pascal-$5^i$.}
	\resizebox{\textwidth}{!}
	{
	\begin{tabular}{c|c|ccccc|ccccc}
		\hline\noalign{\smallskip}
		\multicolumn{1}{c|}{\multirow{2}[0]{*}{Methods}} & \multicolumn{1}{c|}{\multirow{2}[0]{*}{Backbone}} & \multicolumn{5}{c|}{1-shot}  & \multicolumn{5}{c}{5-shot} \\
		\multicolumn{1}{c|}{} & \multicolumn{1}{c|}{} & \multicolumn{1}{c}{f-0} & \multicolumn{1}{c}{f-1} & \multicolumn{1}{c}{f-2} & \multicolumn{1}{c}{f-3} & \multicolumn{1}{c|}{mean} & \multicolumn{1}{c}{f-0} & \multicolumn{1}{c}{f-1} & \multicolumn{1}{c}{f-2} & \multicolumn{1}{c}{f-3} & \multicolumn{1}{c}{mean} \\
		\noalign{\smallskip}\hline\noalign{\smallskip}
		ASR \cite{21-liu2021anti} & \multirow{3}[0]{*}{VGG-16} & 50.2  & 66.4  & 54.3  & 51.8  & 55.7  & 53.7  & 68.5  & 55.0    & 54.8  & 58.0 \\
		PFENet \cite{12-tian2020prior} &  & 56.9  & 68.2  & 54.4  & 52.4  & 58.0    & 59.0    & 69.1  & 54.8  & 52.9  & 59.0 \\
		Ours+PFENet &  & 60.2  & 69.5  & 58.7  & 54.8  & 60.8  & 64.2  & 70.5  & 59.6  & 56.8  & 62.8 \\
		\noalign{\smallskip}\hline\noalign{\smallskip}
		SAGNN \cite{22-xie2021scale} & \multirow{12}[0]{*}{Resnet-50} & 64.7  & 69.6  & 57.0    & 57.2  & 62.1  & 64.9  & 70.0    & 57.0    & 59.3  & 62.8 \\
		ASR \cite{21-liu2021anti} &  & 55.2  & 70.4  & 53.4  & 53.7  & 58.2  & 59.4  & 71.9  & 56.9  & 55.7  & 61.0 \\
		PPNet \cite{17-liu2020part} &  & 47.8  & 58.8  & 53.8  & 45.6  & 51.5  & 58.4  & 67.8  & 64.9  & 56.7  & 62.0 \\
		CMN \cite{19-xie2021few} &  & 64.3  & 70.0    & 57.4  & \textbf{59.4}  & 62.8  & 65.8  & 70.4  & 57.6  & 60.8  & 63.7 \\
		CWT \cite{16-lu2021simpler} &  & 56.3  & 62.0    & 59.9  & 47.2  & 56.4  & 61.3  & 68.5  & 68.5  & 56.6  & 63.7 \\
		MM-Net \cite{15-wu2021learning} &  & 62.7  & 70.2  & 57.3  & 57.0    & 61.8  & 62.2  & 71.5  & 57.5  & 62.4  & 63.4 \\
		ASGNet \cite{14-li2021adaptive} & & 58.8  & 67.9  & 56.8  & 53.7  & 59.3  & 63.7  & 70.6  & 64.2  & 57.4  & 64.0 \\
		SCL \cite{13-zhang2021self} & & 63.0    & 70.0    & 56.5  & 57.7  & 61.8  & 64.5  & 70.9  & 57.3  & 58.7  & 62.9 \\
		CyCTR \cite{23-zhang2021few} &  & 67.8  & 72.8  & 58.0    & 58.0    & 64.2  & 71.1  & 73.2  & 60.5  & 57.5  & 65.6 \\
		PFENet \cite{12-tian2020prior} &  & 61.7  & 69.5  & 55.4  & 56.3  & 60.8  & 63.1  & 70.7  & 55.8  & 57.9  & 61.9 \\
		Ours+PFENet &  & 63.3  & 70.6  & 64.6  & 57.7  & 64.1  & 65.5  & 71.8  & 68.3  & \textbf{62.6}  & 67.1 \\
		Ours+CyCTR &  & 67.9  & \textbf{73.0}    & 61.8  & 58.4  & 65.3  & 69.4  & 73.7  & 61.6  & 60.8  & 66.4 \\
		\noalign{\smallskip}\hline\noalign{\smallskip}
		DAN \cite{18-wang2020few} & \multirow{7}[0]{*}{Resnet-101} & 54.7  & 68.6  & 57.8  & 51.6  & 58.2  & 57.9  & 69.0    & 60.1  & 54.9  & 60.5 \\
		CWT \cite{16-lu2021simpler} &  & 56.9  & 65.2  & 61.2  & 48.8  & 58.0    & 62.6  & 70.2  & 68.8  & 57.2  & 64.7 \\
		ASGNet \cite{14-li2021adaptive} &  & 59.8  & 67.4  & 55.6  & 54.4  & 59.3  & 64.6  & 71.3  & 64.2  & 57.3  & 64.4 \\
		CyCTR \cite{23-zhang2021few} &  & 69.3  & 72.7  & 56.5  & 58.6  & 64.3  & \textbf{73.5}  & \textbf{74.0}    & 58.6  & 60.2  & 66.6 \\
		PFENet \cite{12-tian2020prior} &  & 60.5  & 69.4  & 54.4  & 55.9  & 60.1  & 62.8  & 70.4  & 54.9  & 57.6  & 61.4 \\
		Ours+PFENet &  & 61.8  & 69.7  & \textbf{65.2}  & 57.8  & 63.6  & 64.5  & 71.6  & \textbf{68.6}  & 62.1  & 66.7 \\
		Ours+CyCTR &  & \textbf{70.4}  & 72.5  & 62.4  & 57.9  & \textbf{65.8}  & 72.2  & 73.2  & 63.8  & 60    & \textbf{67.3} \\
		\noalign{\smallskip}\hline
	\end{tabular}
	}
	\label{tab:pascal_miou}%
\end{table}%

\begin{table}[htb]
	\centering
	\caption{Comparison with other state-of-the-arts using mIoU(\%) for 1-shot and 5-shot settings on COCO-$20^i$. CyCTR* is retesting on COCO2014 using the release code provided by authors, where the given results in \cite{23-zhang2021few} are on COCO2017.}
	\resizebox{\textwidth}{!}{
	\begin{tabular}{c|c|ccccc|ccccc}
		\hline\noalign{\smallskip}
		\multicolumn{1}{c|}{\multirow{2}[0]{*}{Methods}} & \multicolumn{1}{c|}{\multirow{2}[0]{*}{Backbone}} & \multicolumn{5}{c|}{1-shot}  & \multicolumn{5}{c}{5-shot} \\
		\multicolumn{1}{c|}{} & \multicolumn{1}{c|}{} & \multicolumn{1}{c}{~~f-0~~} & \multicolumn{1}{c}{~~f-1~~} & \multicolumn{1}{c}{~~f-2~~} & \multicolumn{1}{c}{~~f-3~~} & \multicolumn{1}{c|}{mean} & \multicolumn{1}{c}{~~f-0~~} & \multicolumn{1}{c}{~~f-1~~} & \multicolumn{1}{c}{~~f-2~~} & \multicolumn{1}{c}{~~f-3~~} & \multicolumn{1}{c}{mean} \\
		\noalign{\smallskip}\hline\noalign{\smallskip}
		ASR \cite{21-liu2021anti} & \multirow{9}[0]{*}{ResNet-50} & 30.6  & 36.7  & 32.7  & 35.4  & 33.9  & 33.1  & 39.5  & 34.2  & 36.2  & 35.8 \\
		PPNet \cite{17-liu2020part} &  & 34.5  & 25.4  & 24.3  & 18.6  & 25.7  & \textbf{48.3}  & 30.9  & 35.7  & 30.2  & 36.2 \\
		CMN \cite{19-xie2021few} &  & \textbf{37.9}  & \textbf{44.8}  & 38.7  & 35.6  & 39.3  & 42.0    & \textbf{50.5}  & 41.0    & 38.9  & 43.1 \\
		CWT \cite{16-lu2021simpler} &  & 32.2  & 36.0    & 31.6  & 31.6  & 32.9  & 40.1  & 43.8  & 39.0    & 42.4  & 41.3 \\
		MM-Net \cite{15-wu2021learning} &  & 34.9  & 41.0    & 37.2  & 37.0    & 37.5  & 37.0    & 40.3  & 39.3  & 36.0    & 38.2 \\
		ASGNet \cite{14-li2021adaptive} &  & - & - & - & - & 34.6  & - & - & - & - & 42.5 \\
		CyCTR* \cite{23-zhang2021few} &  & 36.8  & 40.2    & 38.1  & 36.1  & 37.8  & 39.6  & 43.5  & 40.7  & 40.6    & 41.1 \\
		Ours+PFENet &  & 37.2  & 43.6  & \textbf{40.9}  & \textbf{39.1}  & \textbf{40.2}  & 41.7  & 50.4  & \textbf{47.1}  & \textbf{44.5}  & \textbf{45.9} \\
		Ours+CyCTR &  & 36.9     & 40.9     & 39.2     & 39.1     & 39.0     & 41.4     & 42.9     & 43.3     & 43.5     & 42.8 \\
		\noalign{\smallskip}\hline\noalign{\smallskip}
		SAGNN \cite{22-xie2021scale} & \multirow{5}[0]{*}{ResNet-101} & 36.1  & 41.0    & 38.2  & 33.5  & 37.2  & 40.9  & 48.3  & 42.6  & 38.9  & 42.7 \\
		DAN \cite{18-wang2020few} &  & - & - & - & - & 24.4  & - & - & - & - & 29.6 \\
		SCL \cite{13-zhang2021self} &  & 36.4  & 38.6  & 37.5  & 35.4  & 37.0    & 38.9  & 40.5  & 41.5  & 38.7  & 39.9 \\
		PFENet \cite{12-tian2020prior} &  & 34.3  & 33.0    & 32.3  & 30.1  & 32.4  & 38.5  & 38.6  & 38.2  & 34.3  & 37.4 \\
		Ours+PFENet &  & 36.0    & 41.7  & 39.3  & 37.1  & 38.5  & 40.8  & 47.8  & 44.5  & 41.6  & 43.7 \\
		\noalign{\smallskip}\hline
	\end{tabular}
	}
	\label{tab:coco_miou}%
\end{table}%

\begin{table}[!htb]
	\begin{minipage}{0.50\linewidth}
		\centering
		\caption{Comparison with other approaches using FB-IoU(\%) on Pascal-$5^i$.}
		\resizebox{\textwidth}{!}{
		\begin{tabular}{c|c|cc}
			\hline\noalign{\smallskip}
			Methods & Backbone & \multicolumn{1}{c}{1-shot} & \multicolumn{1}{c}{5-shot} \\
			\noalign{\smallskip}\hline\noalign{\smallskip}
			ASR \cite{21-liu2021anti} & \multirow{7}[0]{*}{Resnet-50} & 72.9  & 74.1 \\
			CMN \cite{19-xie2021few} &  & 72.3  & 72.8 \\
			ASGNet \cite{14-li2021adaptive} &  & 69.2  & 74.2 \\
			SCL \cite{13-zhang2021self} &  & 71.9  & 72.8 \\
			PFENet \cite{12-tian2020prior} &  & 73.3  & 73.9 \\
			Ours+PFENet &  & \textbf{74.6}  & \textbf{77.0} \\
			Ours+CyCTR &  & 73.2  & 75.3 \\
			\noalign{\smallskip}\hline\noalign{\smallskip}
			DAN \cite{18-wang2020few} & \multirow{6}[0]{*}{Resnet-101} & 71.9  & 72.3 \\
			ASGNet \cite{14-li2021adaptive} &  & 71.7  & 75.2 \\
			CyCTR \cite{23-zhang2021few} &  & 72.9  & 75.0 \\
			PFENet \cite{12-tian2020prior} &  & 72.9  & 73.5 \\
			Ours+PFENet &  & 74.0    & 76.4 \\
			Ours+CyCTR &  & 73.5  & 74.6 \\
			\noalign{\smallskip}\hline
		\end{tabular}
		}	
		\label{tab:pascal_fbiou}%
	\end{minipage}
	\begin{minipage}{0.50\linewidth}
		\centering
		\caption{Comparison with other approaches using FB-IoU(\%) on COCO-$20^i$.}
		\resizebox{\textwidth}{!}{
		\begin{tabular}{c|c|cc}
			\hline\noalign{\smallskip}
			Methods & Backbone & \multicolumn{1}{c}{1-shot} & \multicolumn{1}{c}{5-shot} \\
			\noalign{\smallskip}\hline\noalign{\smallskip}
			CMN [19] & \multirow{4}[0]{*}{ResNet-50} & 61.7  & 63.3 \\
			ASGNet [14] &  & 60.4  & 67.0 \\
			Ours+PFENet &  & \textbf{64.1}  & \textbf{67.1} \\
			Ours+CyCTR &  & 61.4     & 62.7 \\
			\hline\hline\noalign{\smallskip}
			SAGNN [22] & \multirow{5}[0]{*}{ResNet-101} & 60.9  & 63.4 \\
			DAN [18] &  & 62.3  & 63.9 \\
			PFENet [12] &  & 58.6  & 61.9 \\
			Ours+PFENet &  & 61.6  & 64.7 \\
			\noalign{\smallskip}\hline
		\end{tabular}
		}
		\label{tab:coco_fbiou}%
	\end{minipage}
\end{table}%

\begin{figure*}
	\centering
	\includegraphics[width=0.85\linewidth]{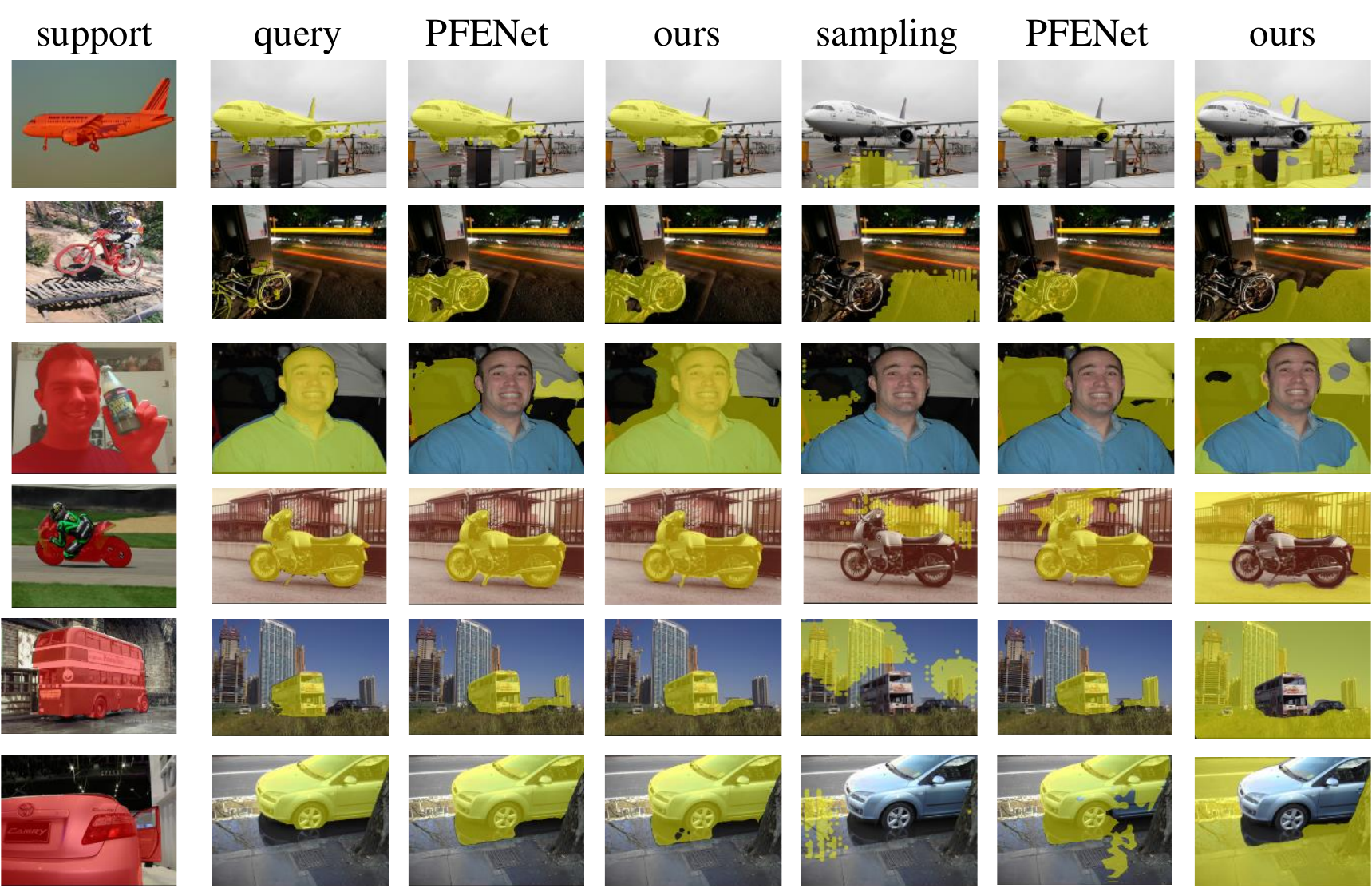}
	\caption{Qualitative results of our method on Pascal-$5^i$. The fifth column is the generated pseudo-mask. Using the prototype of sampling regions, our methods can make correct segmentation, while PFENet still segments the original foreground object.}
	\label{fig:display}
\end{figure*}

\begin{table}[htb]
	\begin{minipage}{0.48\linewidth}
		\centering
		\caption{Ablation studies on the similarity threshold $\delta$ in latent prototype sampling on Pascal-$5^i$.}
		\resizebox{\textwidth}{!}{
		\begin{tabular}{c|cc|cc}
			\hline\noalign{\smallskip}
			\multirow{2}[0]{*}{~$\delta$~} & \multicolumn{2}{c|}{1-shot} & \multicolumn{2}{c}{5-shot} \\
			& \multicolumn{1}{c}{~mIoU~} & \multicolumn{1}{c|}{~FB-IoU~} & \multicolumn{1}{c}{~mIoU~} & \multicolumn{1}{c}{~FB-IoU~} \\
			\noalign{\smallskip}\hline\noalign{\smallskip}
			0.40~   & 63.5  & 72.7  & 65.4  & 73.3 \\
			0.50~   & 63.3  & 73.7  & 66.3  & 75.8 \\
			0.65~  & 64.1  & 74.6  & 67.1  & 77.0 \\
			0.80~   & 64.0    & 74.8  & 66.4  & 76.8 \\
			\noalign{\smallskip}\hline
		\end{tabular}
		}
		\label{tab:ablation1}%
	\end{minipage}
	\begin{minipage}{0.48\linewidth}
		\centering
		\caption{Ablation studies on the weight $\beta$ of contrastive enhancement loss on Pascal-$5^i$.}
		\resizebox{\textwidth}{!}{
		\begin{tabular}{c|cc|cc}
			\hline\noalign{\smallskip}
			\multirow{2}[0]{*}{~$\beta$~} & \multicolumn{2}{c|}{1-shot} & \multicolumn{2}{c}{5-shot} \\
			& \multicolumn{1}{c}{~mIoU~} & \multicolumn{1}{c|}{~FB-IoU~} & \multicolumn{1}{c}{~mIoU~} & \multicolumn{1}{c}{~FB-IoU~} \\
			\noalign{\smallskip}\hline\noalign{\smallskip}
			0.00~     & 62.0    & 72.0    & 63.1  & 72.7 \\
			0.10~   & 64.1  & 74.6  & 67.1  & 77.0 \\
			0.25~  & 63.6  & 74.5  & 67.0    & 76.9 \\
			1.00~     & 59.0    & 71.0    & 62.1  & 72.9 \\
			\noalign{\smallskip}\hline
		\end{tabular}
		}
		\label{tab:ablation2}%
	\end{minipage}
\end{table}%

\begin{table}[htb]
	\centering
	\caption{Ablation studies of the 5-shot settings on Pascal-$5^i$. avg is the setting presented in Section \ref{sec:k-shot}. v-k represents voting is performed on each pixel location, and the location is identified as foreground if voted by k support images.}
	\resizebox{\textwidth}{!}{
	\begin{tabular}{l|rrrrrr|rrrrrr}
		\hline\noalign{\smallskip}
		\multirow{2}[0]{*}{} & \multicolumn{6}{c|}{Pascal-$5^i$}                    & \multicolumn{6}{c}{COCO-$20^i$} \\
		& \multicolumn{1}{c}{~avg~} & \multicolumn{1}{c}{~v-1~} & \multicolumn{1}{c}{~v-2~} & \multicolumn{1}{c}{~v-3~} & \multicolumn{1}{c}{~v-4~} & \multicolumn{1}{c|}{~v-5~~} & \multicolumn{1}{c}{~avg~} & \multicolumn{1}{c}{~v-1~} & \multicolumn{1}{c}{~v-2~} & \multicolumn{1}{c}{~v-3~} & \multicolumn{1}{c}{~v-4~} & \multicolumn{1}{c}{~v-5~~} \\
		\noalign{\smallskip}\hline\noalign{\smallskip}
		~~mIoU~~ & 67.1  & 61.6  & 66.2  & 66.9  & 65.0  & 58.6  & 45.9  & 37.3  & 45.2  & 46.2  & 42.5  & 32.1 \\
		~~FB-IoU~~ & 77.0  & 68.0  & 73.7  & 77.2  & 77.4  & 74.6  & 67.1  & 59.6  & 66.9  & 67.7  & 65.6  & 60.4 \\
		\noalign{\smallskip}\hline
	\end{tabular}
	}
	\label{tab:ablation3}
\end{table}

As shown in Table \ref{tab:pascal_miou}, our approach achieves new state-of-the-art performance by comparisons with other few-shot segmentation approaches on Pascal-$5^i$. In particular, our approach greatly improves the performance of two baselines with different backbones. For ResNet-101, mIoU obtains 3.5\% and 5.3\% improvement on 1-shot and 5-shot task for PFENet, 1.5\% and 0.7\% for CyCTR. It is worth noting that our approach obtains more improvement in 5-shot task. For example, our method significantly outperforms SCL \cite{13-zhang2021self} in 5-shot task, while performance is close in 1-shot task. Table \ref{tab:coco_miou} shows the comparisons with other methods on COCO-$20^i$. Our approach outperforms other approaches in this complex dataset with mIoU increases of 7.8\% (8.5\%) and 1.2\% (1.7\%) respectively for PFENet and CyCTR on 1-shot (5-shot) task. We notice that there exists performance degradation with ResNet-101. The results of 1-shot and 5-shot using FB-IoU evaluation metric are given in Table \ref{tab:pascal_fbiou} and \ref{tab:coco_fbiou}, our approach also achieves state-of-the-art performance.

Some qualitative results on Pascal-$5^i$ are shown in Fig.\ref{fig:display}. Contrastive enhancement can substantially reduce the similarity neglect between prototype and query features. For instance, the baseline mistakenly segments the airplane, although the prototype is generated from the background. Our method can avoid this situation and segment parts of similarity regions. The baseline can not segment the "person" leveraging support images in the third row. There are segmentation mistakes at the object edges, but the enhanced decoder can accurately locate the object. Utilizing the latent objects in the training set, our approach attempts to reduce decoder concentration in query features and drive the decoder to focus more on similarity information.

\subsection{Ablation Study}

The PFENet is chosen as the baseline in ablation study to conduct all experiments. Averaging the evaluations on four splits, we present the results of mIoU and FB-IoU on Pascal-$5^i$.
First, we conduct experiments to study the influence of similarity threshold $\delta$, which controls the similarity of regions belonging to the same category. Table \ref{tab:ablation1} shows that there is slight performance degradation when the threshold decreases because lower threshold introduces noise in pseudo-mask. Additionally, the results are not sensitive to the set thresholds, which means the proposed method is relatively robust. The ablation studies on the weight of contrastive enhancement loss are given in Table \ref{tab:ablation2}. The batchnorm parameters are frozen in our implementation, and unified downsampling is performed in support and query features. Therefore, the performance improves 1.2\% than baseline implementation. When adopting contrastive enhancement loss with weight 0.1, our method increases the mIoU score by 1.8\% and 3.3\% in 1-shot and 5-shot tasks. However, the loss will compel the encoder to leverage too much similarity information with a large weight, harming the model generalization. It indicates that existing framework may exploit query features implicitly for high accuracy.

In Table \ref{tab:ablation3}, we compare different settings on 5-shot segmentation task. The first setting is to average the prototypes from K support images for the query. The other is to perform K forward passes to make prediction, and the pixel position is foreground when k support images vote for it. Table \ref{tab:ablation3} shows the performance of voting is sensitive to the value of k. There are increases in FB-IoU when k is set to 3. Taking the average remains stable in two data sets and requires less inference time, which is adopted in our work.  

\section{Conclusions}

This paper proposes a contrastive enhancement method using latent prototypes for few-shot segmentation. Aiming to mine novel categories and intensify the prototype learning architecture, we design two modules, called Latent Prototype Sampling (LPS) and Contrastive Enhancement (CE). The LPS module leverages feature similarity to sample image regions of the same category and generates the latent prototype. Additionally, the CE module utilizes the latent prototype to make the model focus more on similar information between prototype and query feature for prediction. CE is an auxiliary path without additional parameters. Extensive experiments demonstrate our approach achieve state-of-the-art performance on Pascal-$5^i$ and COCO-$20^i$. In the future, we will explore the application of our methods on multiple prototype learning approaches.

\clearpage
\bibliographystyle{splncs04}
\bibliography{egbib}

\end{document}